# Deep generative models in DataSHIELD

S Lenz[1*], H Binder[1]

[1]Institute of Medical Biometry and Statistics, Faculty of Medicine and Medical Center – University of Freiburg, Freiburg

[*]Corresponding author (E-mail: lenz@imbi.uni-freiburg.de)

## Abstract

The best way to calculate statistics from medical data is to use the data of individual patients. In some settings, this data is difficult to obtain due to privacy restrictions. In Germany, for example, it is not possible to pool routine data from different hospitals for research purposes without the consent of the patients. The DataSHIELD software provides an infrastructure and a set of statistical methods for joint analyses of distributed data. The contained algorithms are reformulated to work with aggregated data from the participating sites instead of the individual data. If a desired algorithm is not implemented in DataSHIELD or cannot be reformulated in such a way, using artificial data is an alternative. We present a methodology together with a software implementation that builds on DataSHIELD to create artificial data that preserve complex patterns from distributed individual patient data. Such data sets of artificial patients, which are not linked to real patients, can then be used for joint analyses. Models that are able to capture a joint distribution of variables and can generate new data according to this distribution are called generative models. Here we focus on deep Boltzmann machines (DBMs), a generative model trained with unsupervised deep learning. Algorithms for training and evaluating DBMs are implemented in the package "BoltzmannMachines" for the Julia programming language. We wrap its functionality for use with DataSHIELD, which is based on the R programming language. Like DataSHIELD, our software is freely available as open source software. As an exemplary application, we conduct a distributed analysis with DBMs on a synthetic data set. The data set simulates genetic variant data. Hidden in noise, it contains patterns of groups of specific mutations that could be linked to some hypothetical pathological outcome. The data is distributed among different sites and a joint artificial data set is generated by DBMs that are trained on each share of the data. The patterns can be recovered in the artificial data using hierarchical clustering of the virtual patients, which demonstrates the feasibility of the approach. Our implementation adds to DataSHIELD the ability to generate artificial data that can be used for various analyses, e. g. for pattern recognition with deep learning. This also demonstrates more generally how DataSHIELD can be flexibly extended with advanced algorithms from languages other than R.

# Introduction

In large consortia, pooling of individual level data is often not possible due to data security and data protection concerns. Thus, techniques for distributed privacy-preserving analysis are needed. For example the MIRACUM consortium [1], a joint project of ten university hospitals in Germany, aims to show how patient data that are distributed across sites can be jointly analysed. In this consortium, a particular goal is to apply advanced machine learning techniques that can find complex interaction patterns in medical data.

One general way to enable such analysis techniques on distributed data is to use a synthetic data approach. Synthetic datasets mimic statistical features of the original data without any linkage to individuals in the original data. These synthetic data can then be shared across the sites for joint analyses. For simple statistical analyses, this approach has been found to work well [2–4], and there are even commercial offerings for business data [5]. Multivariable statistical analyses are also feasible. For example, an approach using bivariate copulas can recreate complex marginal distributions and provide results similar to the original data when using multivariable linear mixed regression for analysis [6].

However, it is still unclear how best to create synthetic data that also reflect complex patterns, which might, e.g., be analysed using machine learning tools. This may require more complex approaches to generate the synthetic data. In particular, generative deep learning approaches might be useful, as they can represent complex patterns [7] and have been shown to be feasible for small sample sizes [8]. Correspondingly, we decided to develop an implementation for artificial data based on deep learning within the DataSHIELD framework for distributed analysis. DataSHIELD [9] is a software tool used in many multicentre studies for distributed privacy-preserving analysis, and which offers many statistical tools for researchers. Its implementation is based on meta-analysis techniques or parameter estimation via distributed calculation. A synthetic data approach in DataSHIELD will thus provide even more flexible data analysis tools to an already very active use community.

In the following, we present the implementation of our approach using deep Boltzmann machines as generative models in DataSHIELD. We used the Julia programming language [10], which is better suited for implementing deep learning algorithms, and integrated it in the statistical analysis environment R [11], which is the basis for DataSHIELD, via a package for interfacing Julia and R. Deep Boltzmann machines were chosen as generative models for synthetic data on the basis of their good performance on data sets with small sample sizes [8]. This is of particular importance, e.g., when the overall sample size is moderate, but sample size per site is small. We demonstrate the feasibility of the approach with an empirical study investigating different numbers of sites and sample sizes per site in a distributed analysis.

# Methods

## DataSHIELD

DataSHIELD is open-source software that is already used in the field of epidemiology for the analysis of multi-centre cohort studies. Analyses in DataSHIELD are performed without individual data leaving the sites. This is possible by using reformulated algorithms that solely rely on aggregated statistics. Only those aggregated statistics leave the sites and are used to calculate the final result. In this way, the DataSHIELD software allows users to perform several types of descriptive statistics and standard statistical models. For example, it is possible to compute linear regression models via DataSHIELD on data sets that are distributed among several sites, and get the same results as with pooled data. The user can access the DataSHIELD functionality by using functions of specific packages in the R programming language [11].

The Opal web server software [12], running in separate instances at each of the sites participating in a federated analysis, provides the decentralised data. Its interface is secured by authentication and authorization. Some users may have the right to view data, while others may only access aggregated data by calling specific R functions that are approved by the organization operating the Opal instance. These R functions, most of which are collected in packages specifically for use in DataSHIELD, must only return data that does not disclose information about individuals. The official DataSHIELD packages are designed and reviewed specifically to minimise the disclosure risk. In addition to the existing package ecosystem, the infrastructure is extensible and allows developers to write their own R packages, which then can be installed by administrators of Opal instances.

## Deep Boltzmann machines (DBMs) as generative models

The goal of generative models is to capture the probability distribution of multiple variables in a model, allowing new samples to be drawn from the model according to this distribution. Generative models are trained in an unsupervised manner with data from the original distribution as input. In many cases, they can also be used to find higher-level representations of the data [7] by analysing the model parameters.

Here we will focus on deep Boltzmann machines as generative models. General Boltzmann machines are stochastic neural networks whose nodes have an activation probability $p(v,h)$ that is determined by the energy function $E$ of the network.

$$p(v,h) = \frac{e^{-E(v,h)}}{Z} \quad \text{with} \quad Z = \sum_{v,h} e^{-E(v,h)}$$

Thus, Boltzmann machines are so-called "energy-based" models. The nodes are divided into two groups. The visible nodes ($v$) receive the data input, while the hidden nodes ($h$) encode latent variables of the data. The normalization constant $Z$ is also called the *partition function.* Due to the large number of terms in the sum for $Z$, which runs over all possible configurations of activations of nodes, computing the real value of the probability is too complex for most use cases. In practice, Gibbs sampling is used instead to sample from the model. With Gibbs sampling, it is also easy to sample conditionally on specific variables, which makes it possible to use Boltzmann machines as generative models in "what-if" scenarios. For example, in a medical setting, a Boltzmann machine trained on patients' diagnoses can be used to generate synthetic patient data with specific disease patterns, even if these patterns are relatively rare in the original data. A use case for this may be to simulate a population of patients in planning a new study.

The network of general Boltzmann machines is a complete undirected graph, where all nodes are connected to each other. A first step in making Boltzmann machines practically usable was to use *restricted Boltzmann machines* (RBMs). These restrict the connections in the graph, disallowing connections between visible nodes as well as connections between hidden nodes. Thereby, the graph of the network forms a complete bipartite graph that partitions the set of vertices into the set of visible nodes, the *visible layer*, and the set of hidden nodes, the *hidden layer*. This allows for the rapid calculation of the conditional probabilities because in this case it is possible to derive simple formulas for the conditional probabilities, which can be calculated for all nodes in a layer in a vectorised way. This is the basis for an effective training algorithm for RBMs called *contrastive divergence*[13]. RBMs can also be used as generative models but the restrictions on the connections in the network also limit their power to model distributions.

Subsequently, it was discovered that stacking restricted Boltzmann machines on top of each other by training the next restricted Boltzmann machines with the hidden activations of the previous one enabled the networks to learn features of increasing abstraction. The resulting model is a *deep belief network* (DBN), and the training procedure is called greedy layer-wise training [14]. Although this architecture is well suited for dimension reduction, it is less powerful than the next stage of development, the *deep Boltzmann machines* (DBMs). Deep belief networks can be used as generative models by sampling in the last restricted Boltzmann machine and then using the conditional probabilities to propagate the activation to the visible nodes. This way, the full information of the network is not harnessed equally to generate new samples. Deep Boltzmann machines, on the other hand, have the same network layout as deep belief networks, but generate samples employing the full network. This is similar to a general Boltzmann machine, albeit restricted to a layered layout. For training, DBMs are optimized with an algorithm for maximizing the variational lower bound of the likelihood in the Boltzmann machine model. This algorithm is also referred to as *fine-tuning* because greedy layer-wise pre-training is carried out to provide a good starting point for the variational likelihood algorithm, which would otherwise not succeed in finding a good local optimum.

Due to the intractable nature of the partition function, monitoring the optimisation process and evaluating the resulting model is difficult. For restricted Boltzmann machines, the *reconstruction error* is a proxy measure that follows the likelihood of the model very well. It is calculated by taking the data as activation of the visible nodes, calculating the probabilities for the activations in the hidden layer conditioned on the data input and then again calculating the probabilities of the visible layer conditioned on the hidden probabilities. This last result is called the reconstruction of the data input. The reconstruction error is the distance to the original data. It can also be used to monitor the greedy layer-wise training of deep belief networks and for the pre-training of deep Boltzmann machines.

With a stochastic algorithm called *annealed importance sampling* (AIS), it is possible to estimate the likelihood of restricted and deep Boltzmann machines [15]. It is needed in particular to measure the training objective of DBMs, the variational lower bound of the likelihood.

We employ a package for the Julia programming language [10] that implements these algorithms and provides a user-friendly interface for training and evaluating deep Boltzmann machines (https://github.com/stefan-m-lenz/BoltzmannMachines.jl). Further on, we show how we integrate this with the DataSHIELD software and concept.

## Implementation of deep Boltzmann machines in DataSHIELD

The developed software allows users to remotely train deep Boltzmann machines without access to the individual-level data. Trained models can be used, e.g., to generate synthetic data as depicted in Fig 1.

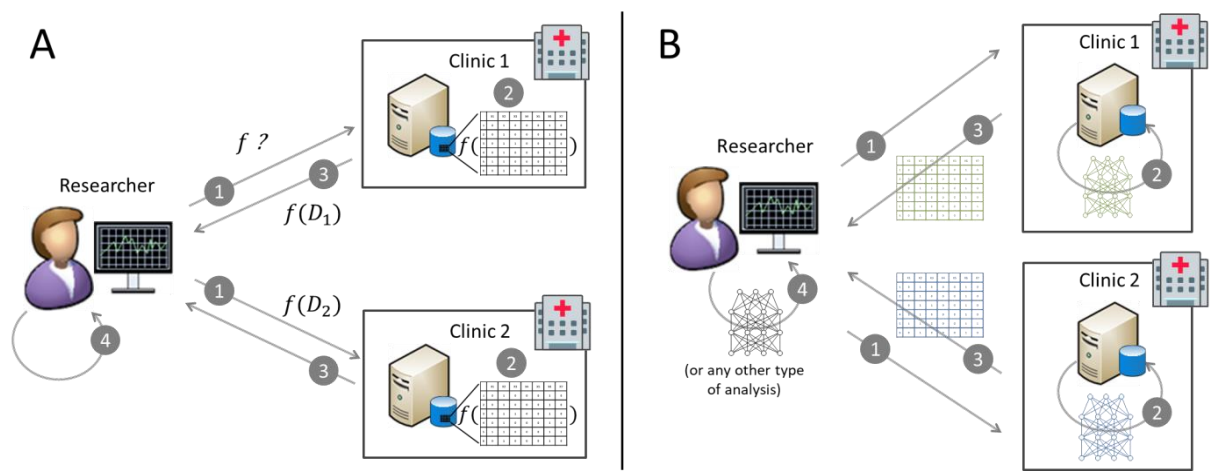

**Fig 1. Applying the DataSHIELD principle in working with synthetic data from generative models.** The standard DataSHIELD approach is depicted in panel A: The researcher sends a request via the DataSHIELD infrastructure (1). The sites then calculate aggregated statistics (2) and return them to the researcher (3). These statistics do not allow conclusions about individual patients, but can be used to derive useful information about the population (4). When working with generated models and synthetic data (panel B), the workflow is similar. The researcher requests the training of a generative model (1). Once the model has been trained (2), synthetic samples can be generated (3). The researcher can use the synthetic data to conduct further analyses (4).

We provide our implementation of DBMs in DataSHIELD as open-source software. It consists of a client-side R package (https://github.com/stefan-m-lenz/dsBoltzmannMachines) and a server-side R package that can be installed in an Opal server and called from the client-side (https://github.com/stefan-m-lenz/dsBoltzmannMachinesClient). The server-side needs Julia with the package "BoltzmannMachines" installed in addition. The functionality of the "BoltzmannMachines" Julia package is imported into R via the "JuliaConnectoR" R package (https://github.com/stefan-m-lenz/JuliaConnectoR). This provides a generic interface that allows the use of Julia functions in R and thereby makes it possible to obtain the speed advantages of the Julia code while using the DataSHIELD R interface.

From a technical perspective, it is straightforward to transfer the models outside via DataSHIELD, because the DataSHIELD infrastructure can transfer arbitrary R objects to the client. It is the responsibility of the developers of DataSHIELD functionality to ensure that the returned values do not disclose sensitive information about individuals. For basic aggregated statistics it is possible to prove mathematically how much information about individuals is contained. For neural networks, this is very hard because they are so complex and usually consist of a very high number of parameters. In many cases, the number of parameters is even higher than the number of data points in the training data itself. Thus, it is very hard to prove that a neural network cannot be hacked. Model inversion attacks, which aim to extract information about individual data sets from trained models, are being researched and developed [16]. Therefore, we do not allow the transfer of the models by default, but give data custodians the option to explicitly allow this in the Opal server environment if there is enough trust in the given setting.

An additional challenge, common to all neural networks, is the extensive hyperparameter tuning that the training requires. As shown in Fig 2, the number of epochs and the learning rate, together with the model architecture, are parameters that are highly important for successful training. These parameters must be tuned individually for different data sets, as the learning rate depends on how informative the different samples are, and the number of epochs must be adjusted accordingly. The architecture must be deep enough to be able to capture the important structure. At the same time the model should not have too many parameters to avoid overfitting and computational cost. To choose these parameters, our software provides different metrics to assess the model quality during and after the training. It offers functions to estimate the likelihood (for RBMs and DBMs) and the lower bound of the likelihood (for DBMs) via AIS. For smaller models, it is also possible to calculate the likelihood exactly. These evaluations can be collected during training to monitor its success. The monitoring output can be transferred and displayed to the DataSHIELD client without privacy issues, even if the number of training attempts is high, because it does not contain information about individual patient data. In this way, the user can see the performance and select good hyperparameters without having direct access to the models. After a successful training, the final model can then be used to generate synthetic data that is handed to the researcher.

**Table 1 Overview of client-side functions for training and using DBM models**

| Function name | Short description |
|---|---|
| ds.monitored_fitrbm | Monitored training of an RBM model |
| ds.monitored_stackrbms | Monitored training of a stack of RBMs. Can be used for pre-training a DBM or for training a DBN |
| ds.monitored_fitdbm | Monitored training of a DBM, including pre-training and fine-tuning |
| ds.setJuliaSeed | Set a seed for the random number generator |
| ds.dbm.samples/ ds.rbm.samples | Generate samples from a DBM/RBM. This also allows conditional sampling. |
| ds.bm.defineLayer | Define training parameters individually for a RBM layer in a DBM or DBN |
| ds.bm.definePartitionedLayer | Define a partitioned layer using other layers as parts |
| ds.dbm.top2LatentDims | Get a two-dimensional representation of latent features |
| ds.rbm.loglikelihood | Estimates the partition function of an RBM with AIS and then calculates the log-likelihood |
| ds.dbm.loglikelihood | Performs a separate AIS run for each of the samples to estimate the log-likelihood of a DBM |
| ds.dbm.logproblowerbound | Estimates the variational lower bound of the likelihood of a DBM with AIS |
| ds.rbm.exactloglikelihood/ ds.dbm.exactloglikelihood | Calculates the log-likelihood for a RBM/DBM (exponential complexity) |

```r
library(dsBoltzmannMachinesClient)
logindata <- data.frame(server = "server",
                        url = "https://datashield.example.com",
                        user = "user", password = "password",
                        table = "MyTable")

o <- datashield.login(logins = logindata, assign = TRUE) # ==> data in D

ds.setJuliaSeed(o, 1) # for reproducibility

ds.splitdata(o, "D", 0.2, "D.Train", "D.Test")

# Training
result <- ds.monitored_fitdbm(o, data = "D.Train", nhiddens = c(50, 25, 15),
                              epochspretraining = 30,
                              learningratepretraining = 0.005,
                              epochs = 100,
                              learningrate = 0.05,
                              monitoringdata = c("D.Train", "D.Test"))
plotMonitoring(result)

# Generating data
generated <- ds.dbm.samples(o)

datashield.logout(o)
```

**Fig 2 Example code for training a deep Boltzmann machine and using it as a generative model.** First, the user needs to log in to the Opal server, where the data is stored. If the specified data set is available, and the user has the correct access rights, the data set is loaded into the R session. The loaded data can be split into training and test data before the training. In the subsequent call to the fitting function, which by default also collects monitoring data from the training, the most important parameters for training a DBM are included. The numbers of hidden nodes for each of the hidden layers ("nhiddens") determine the model architecture. The learning rate and the number of epochs for pre-training and fine-tuning of the DBM are the most important parameters for the optimization procedure. If a good solution has been found, the model can be used to generate synthetic data and return it to the client.

To provide a rough idea of the time needed to train DBMs with our software, we measured the execution time of the code in Fig 2, using the example data set described above, on a desktop computer with an Intel Core i5-4570 processor with 3.2 GHz, and running Opal in a virtual machine with only 1 CPU. The fitting of the DBM with monitoring took less than 1.5 minutes, and without monitoring less than 15 seconds. (On the first training run in the session, the elapsed times were a little longer due to compilation in Julia.)

# Results

We consider an example that is motivated by genetic variant data, so-called SNPs (single nucleotide polymorphisms). One main goal of deep learning on genetic data is to uncover interactions between genetic mutations that lead to certain (pathological) phenotypes. Particularly interesting are cases with many interacting mutations that are jointly responsible for a resulting phenotype. These cases are hard to detect with univariate testing of SNPs, which is used in genome-wide association studies (GWAS).

We conducted an experiment to show that is it possible to learn and reproduce higher-level patterns employing DBMs in SNP-like data sets and evaluated their performance as generative models in this setting, showing the effect of the available sample size in particular (see Fig 3). The artificial data set for our experiment consisted of binary data mostly consisting of zeros with some noise added from a Bernoulli distribution with a probability of 0.1. The 500 samples are split equally into "cases" and "controls". The "cases" have groups of five ones at five possible "SNP sets" among the 50 SNP variables. This could correspond to mutations that may deactivate a certain pathway, when they occur together.

In this experiment the data set is split onto a number of virtual sites, where models are trained and then used to generate new data. This new data can then be visually compared with the original data (see Fig 3).

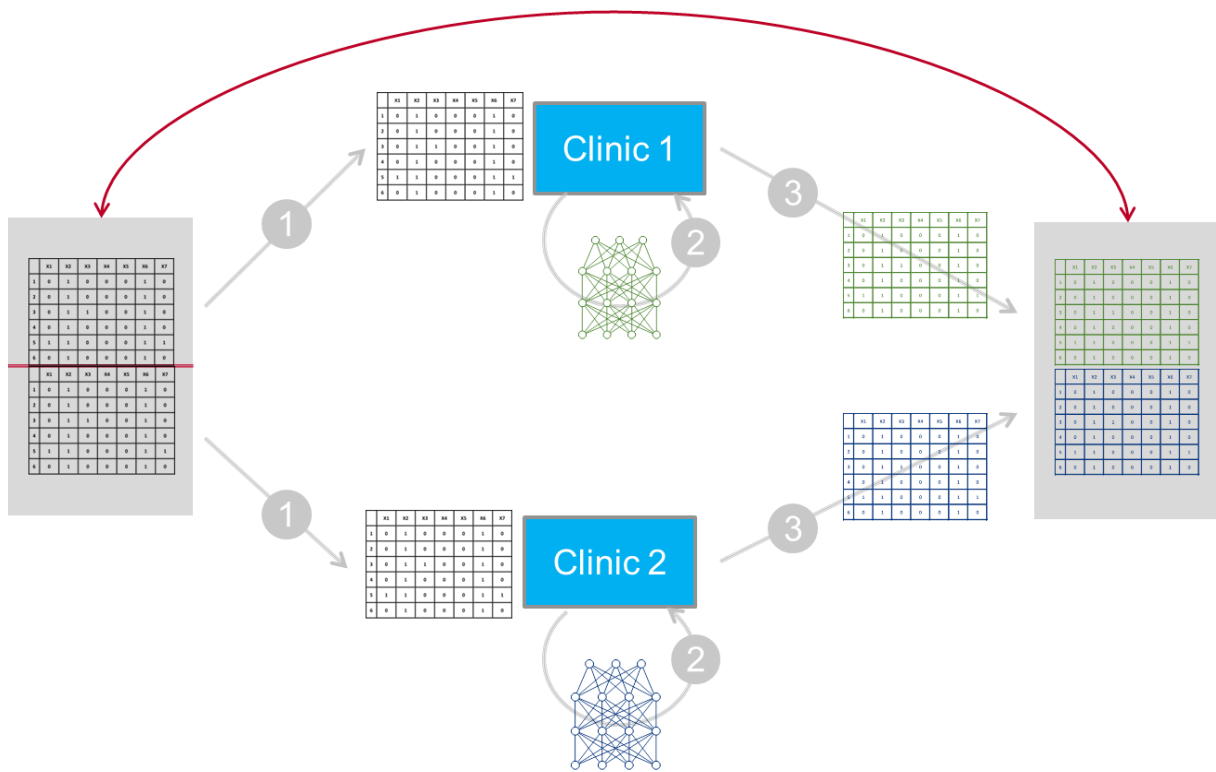

**Fig 3. Sketch of the experimental setup for the comparison of original and generated data.** In the first step, the original data is split into a number of smaller data sets, which are distributed to the virtual sites. (For simplicity, only two sites/clinics are shown.) In Step 2, separate generative models are trained at each site on their share of the data. In step 3, synthetic data are generated by each of the models and compiled to again form one overall data set. This synthetic data set will be visually compared to the original data set. For the results, see Fig 4 below.

The results of the experiment are shown in Fig 4. The higher-level patterns, which here are the arrays of co-occurring SNPs, are preserved in the synthetic data, even in the case of 20 sites having only 25 patients. However, one can observe that the noise in the sampling output increases as the same amount of samples is distributed among a growing number of sites. Further, it is notable that there is some price to pay for using synthetic data since the output is not exactly the same as the input (see Fig 4, e.g. comparing panels A and B).

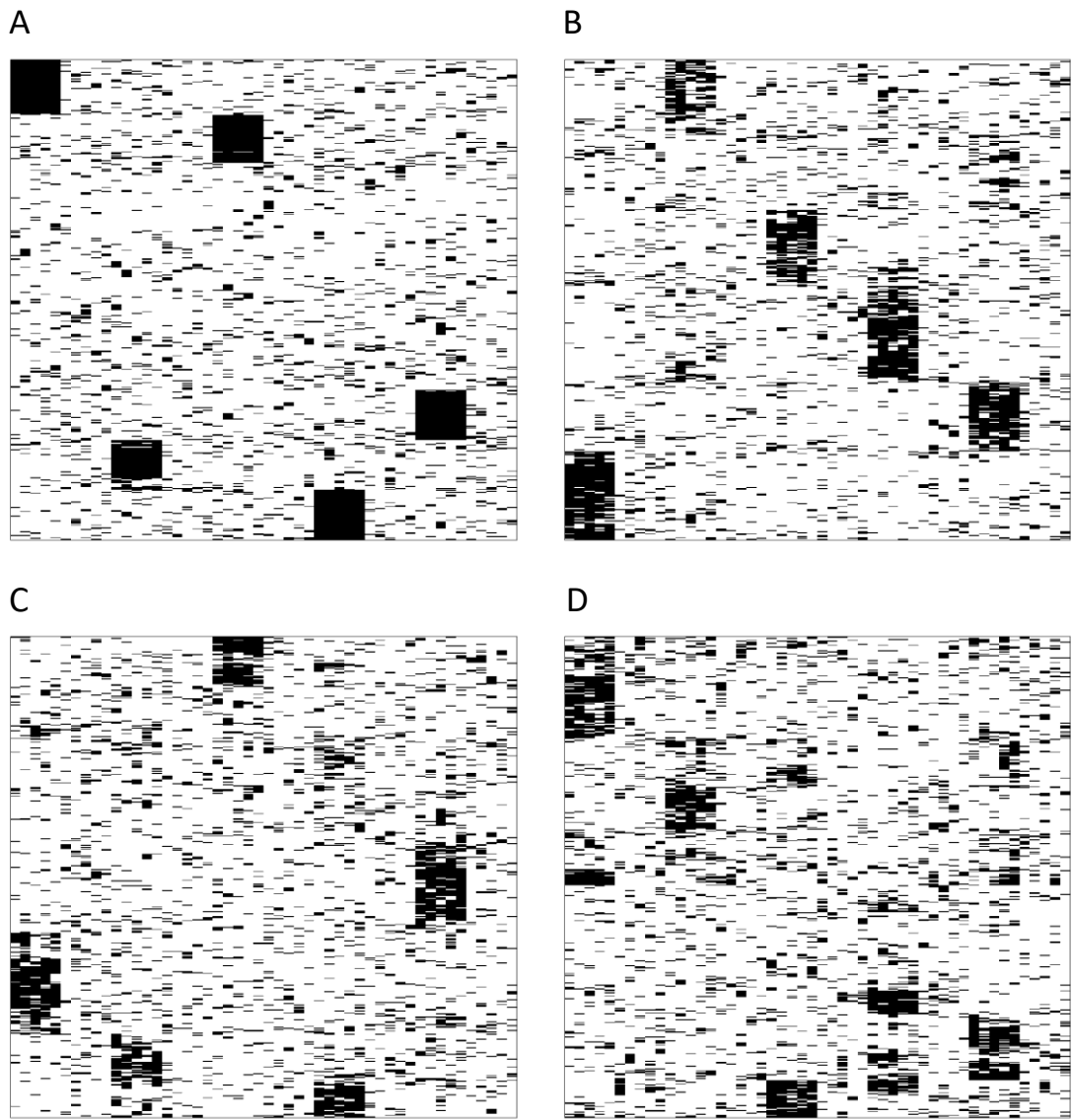

**Fig 4. Hierarchical clustering view of a data set and associated synthetic data sets.** The rows are the patients and the columns are the variables. The rows are clustered hierarchically [17]. Panel A shows the original data set, panel B shows data generated from one DBM that has been trained on the original data. Panels C and D show outputs of the experiment conducted with 2 and 20 sites, respectively.

# Discussion

While synthetic data are a promising option for enabling a broad set of statistical analyses in a distributed setting, an accessible implementation is currently lacking. We described an extension of the popular DataSHIELD framework for distributed analysis under data protection constraints. In particular, we leveraged generative deep learning for obtaining synthetic data, connecting the R language, which is the basis for DataSHIELD, to the Julia language. To the user, this complexity is masked via a convenient R package.

We chose deep Boltzmann machines (DBMs) as generative models to be implemented in DataSHIELD because of their advantages in certain use cases. As shown in our feasibility study, DBMs can deal with a low number of training samples and are hard to overfit. This makes them especially suitable for distributed settings with small data sets at each of the sites. In recent years, other generative models, most importantly generative adversarial networks (GANs) [18] and variational autoencoders (VAEs) [19,20], have become popular. In contrast to deep Boltzmann machines, which rely on Markov Chain Monte Carlo methods to be trained, these models are trained using backpropagation of errors, requiring a different implementation approach for the training. As a next step, we plan to implement VAEs as a generative model in DataSHIELD, and to combine the different models to minimize bias in generated data.

Another area in which DBMs excel is conditional sampling. A possible application of conditional sampling is a DBM trained on gene expression data, to simulate the up-regulation of one pathway and observe changes in the expression of other genes. Another example could be to generate data for medication or comorbidities conditioned on diagnoses in data from electronic health records. In DBMs, conditional sampling is straightforward using Gibbs sampling.

The results of our empirical investigation showed that some structure is maintained even with very small sample sizes, but performance could potentially still be improved. For example, partitioning of layers can be used to further decrease the number of samples needed to find informative structure in the data [8].

## Conclusions

With the presented extension to the DataSHIELD software, we add the possibility of generating artificial data sets that preserve the higher level patterns from individual patient data that may be distributed among different sites. These generated data sets can then be analysed to extract patterns from the original data without access to individual patient data. The results presented here indicate that our proposed approach is ready for use in real world applications. This is facilitated by the user-friendly design of our implementation complemented by extensive documentation. More generally, the proposed implementation provides a sound basis for subsequent extensions to other generative approaches for synthetic data in distributed analysis.

# Conflicts of Interest

None declared.

# Acknowledgements

This work has been supported by the Federal Ministry of Education and Research (BMBF) in Germany in the MIRACUM project (FKZ 01ZZ1801B).